\documentclass{article}

\usepackage{arxiv}
\usepackage{cite}
\usepackage{amsmath,amssymb,amsfonts}
\usepackage{graphicx}
\usepackage{textcomp}
\usepackage{xcolor}
\usepackage[T1]{fontenc}
\usepackage{url}
\usepackage{hyperref}
\usepackage{microtype}
\usepackage{times}

\graphicspath{{./images/}}
\begin{document}

\title{Explainability Assistant: A Conversational XAI Interface for Interpreting Energy Consumption Models}

\author{
  Rodion Krjut\v{s}kov \\
  Nupp Software\\
  Tallinn, Estonia\\
  \texttt{rodion\@nupp.dev} \\
  \And
  Eduard Barbu\,\href{https://orcid.org/0000-0002-3664-5367}{\textnormal{\scriptsize[ORCID]}} \\
  Institute of Computer Science\\
  University of Tartu\\
  Tartu, Estonia\\
  \texttt{eduard.barbu\@ut.ee} \\
  \And
  Nikos Sakkas \\
  Apintech Ltd\\
  POLIS-21 Group\\
  Cyprus\\
  \texttt{sakkas\@apintech.com} \\
  \And
  Sofia Yfanti \\
  Department of Mechanical Engineering\\
  Hellenic Mediterranean University\\
  Heraklion, Greece\\
  \texttt{sifanti\@hmu.gr} \\
}

\maketitle

\begin{abstract}
Energy consumption forecasting relies on increasingly complex machine learning (ML) models, such as Genetic Programming-based symbolic regressors, whose predictions can be difficult for facility managers and building operators to interpret. Explainable Artificial Intelligence (XAI) techniques address this opacity, but traditional XAI dashboards require substantial technical expertise and provide limited flexibility for dynamic, context-aware inquiry. Conversational XAI systems offer a promising alternative; however, previous approaches, such as TalkToModel, were constrained by rigid custom grammars and achieved only 76.8\% intent-parsing accuracy. This paper introduces the \textbf{Explainability Assistant}, an open-source conversational XAI system that leverages the function-calling capabilities of modern Large Language Models (LLMs) to overcome these limitations. The system achieves 94\% intent-parsing accuracy, supports flexible natural language interaction, and adapts to different ML problem types without task-specific fine-tuning. We present the system's architecture and report results from a comparative evaluation conducted with energy domain specialists, contrasting the Explainability Assistant with a traditional XAI dashboard. The evaluation suggests improved usability and consistent task accuracy, with all experts unanimously preferring the conversational interface for practical use.
\end{abstract}


\section{Introduction}

ML models are increasingly deployed in high-stakes domains such as energy management, healthcare, and finance, where understanding model predictions is critical for trust and accountability~\cite{ribeiro2016should,lundberg2017unified}. In building energy systems, for example, facility managers rely on consumption forecasts produced by complex models, yet have limited means to verify why a particular prediction was made. Post-hoc explainability techniques, such as LIME~\cite{ribeiro2016should} and SHAP~\cite{lundberg2017unified}, as well as counterfactual explanations~\cite{mothilal2020explaining}, have emerged as popular tools to address this opacity. However, practitioners struggle to select appropriate explanation methods, interpret their outputs, and answer follow-up questions during model analysis~\cite{kaur2020interpreting}.

Traditional explainability interfaces, such as point-and-click dashboards, require users to possess substantial technical expertise and often fail to support flexible, iterative exploration. This is particularly limiting in applied settings such as building energy management, where the end users are facility managers and engineers rather than data scientists. Natural language dialogue systems offer a promising alternative, enabling open-ended and accessible interactions that allow users to ask questions, explore model behavior dynamically, and receive contextually relevant explanations without needing to understand the underlying technical machinery.

Building on the foundations of TalkToModel~\cite{slack2023talktomodel}, a conversational system that translates user utterances into executable operations for explainability, we introduce the \textbf{Explainability Assistant} (EA), an enhanced dialogue-based system for understanding ML models. The Explainability Assistant addresses key limitations of TalkToModel by leveraging recent advances in LLMs with function-calling capabilities, enabling more accurate natural language understanding, improved architectural modularity, and broader applicability across diverse use cases.

We evaluate the Explainability Assistant in energy consumption forecasting and demonstrate that it achieves significantly higher parsing accuracy (up to 94\%), surpassing the 76.8\% accuracy reported for previous grammar-based approaches. Through evaluation using both automatic metrics and stakeholder feedback, we show that the system supports diverse explainability tasks, including feature importance analysis, counterfactual reasoning, ML model error analysis, and what-if queries.

Our contributions are threefold:
\begin{enumerate}
    \item A conversational architecture that integrates modern LLMs for natural language understanding in explainability contexts.
    \item A lightweight, modular system design that separates concerns and enables flexible deployment.
    \item Empirical evidence that dialogue-based interfaces can support improved model understanding for domain experts and ML practitioners.
\end{enumerate}

The rest of the paper is organized as follows. Section~\ref{sec:related} reviews
relevant work in explainability and conversational interfaces. Section~\ref{sec:system}
presents the architecture and XAI tools underlying the Explainability Assistant.
Section~\ref{sec:evaluation} reports results from the parsing evaluation and the
expert validation with energy domain specialists. Section~\ref{sec:conclusion} concludes the paper.

We provide all materials required to assess the system: a \textbf{live demo}, a \textbf{video walkthrough of the system}, the \textbf{gold parse datasets}, the \textbf{source code} (front-end and back-end), and the
\textbf{full expert validation questionnaire}. All links are collected in Appendix~\ref{sec:resources}.

\section{Related Work}
\label{sec:related}

The need for interpretable ML systems has driven extensive research into explainability techniques. Early work focused on inherently interpretable models~\cite{lou2013accurate,caruana2015intelligible}, while post-hoc methods emerged for black-box models. LIME~\cite{ribeiro2016should} and SHAP~\cite{lundberg2017unified} compute feature importance, counterfactual explanations~\cite{mothilal2020explaining} identify minimal input changes that alter predictions, and gradient-based methods~\cite{selvaraju2017grad} visualize neural network behavior. Despite these advances, practitioners struggle to select appropriate methods and interpret results~\cite{kaur2020interpreting,liao2020questioning}.

Interactive tools combine multiple explanation techniques into unified interfaces. Dashboard-based systems, such as the Language Interpretability Tool~\cite{tenney2020language} and the What-If Tool~\cite{wexler2019if}, provide point-and-click access to explanations, but still require users to understand which operations to perform. TalkToModel~\cite{slack2023talktomodel} introduced natural language dialogue for explainability using fine-tuned language models with custom grammars, demonstrating improved user understanding but achieving only 76.8\% parsing accuracy and requiring complex task-specific adaptation.

Modern LLMs, including the GPT, Llama, and Gemini families, support function-calling mechanisms for tool orchestration, enabling applications in conversational QA~\cite{gao2023pal}, data analysis~\cite{zhang2024datacopilot}, and agentic tasks~\cite{qin2023toolllm,schick2024toolformer}. Recent work has begun applying these capabilities to XAI solutions. Wang et al.~\cite{wang2024llmcheckup} introduced LLMCheckup, a conversational XAI system that uses prompt engineering and in-context learning to enable dialogue-based model explanations, relying on prompt chaining to invoke explanation methods. Samimi et al.~\cite{samimi2025visual} developed a visual-conversational interface for diabetes risk prediction that combines a fine-tuned T5 parser with LLM-based dialogue and interactive visualizations, retaining a task-specific fine-tuned component alongside LLMs. At a different level of abstraction, Shaham et al.~\cite{rottshaham2025maia} proposed MAIA, a multimodal agent that interprets neural network internals through autonomous experimentation, targeting model internals interpretability for researchers rather than prediction-level explainability for end users.

The Explainability Assistant differs from these systems in its use of structured LLM function-calling to translate natural language queries into explainability operations, providing more predictable tool invocation than prompt chaining~\cite{wang2024llmcheckup}, eliminating task-specific fine-tuning requirements~\cite{samimi2025visual}, and focusing on prediction-level explanations accessible to domain experts in applied settings such as energy management.

\section{System Description}
\label{sec:system}

\subsection{System Architecture}

\begin{figure*}
    \centering
    \includegraphics[width=0.8\linewidth]{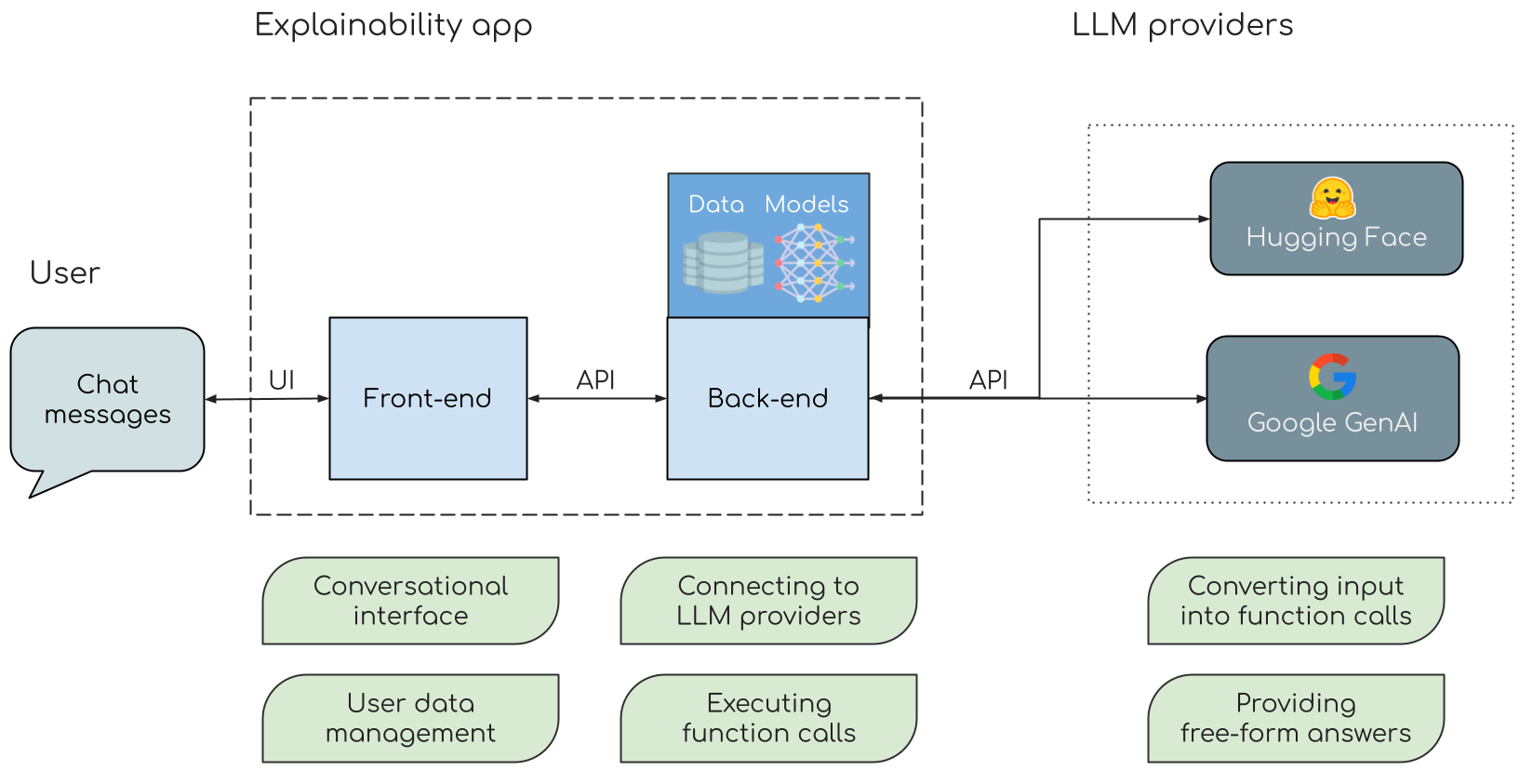}
    \caption{Explainability Assistant system architecture.}
    \label{fig:architecture}
\end{figure*}

The Explainability Assistant employs a modular architecture that decouples the presentation layer from the core computational logic, as illustrated in Figure~\ref{fig:architecture}. This separation enhances maintainability, enables independent scaling of components, and facilitates integration with diverse frameworks and LLM providers.

\begin{figure*}[tbp]
    \centering
    \includegraphics[width=0.77\linewidth]{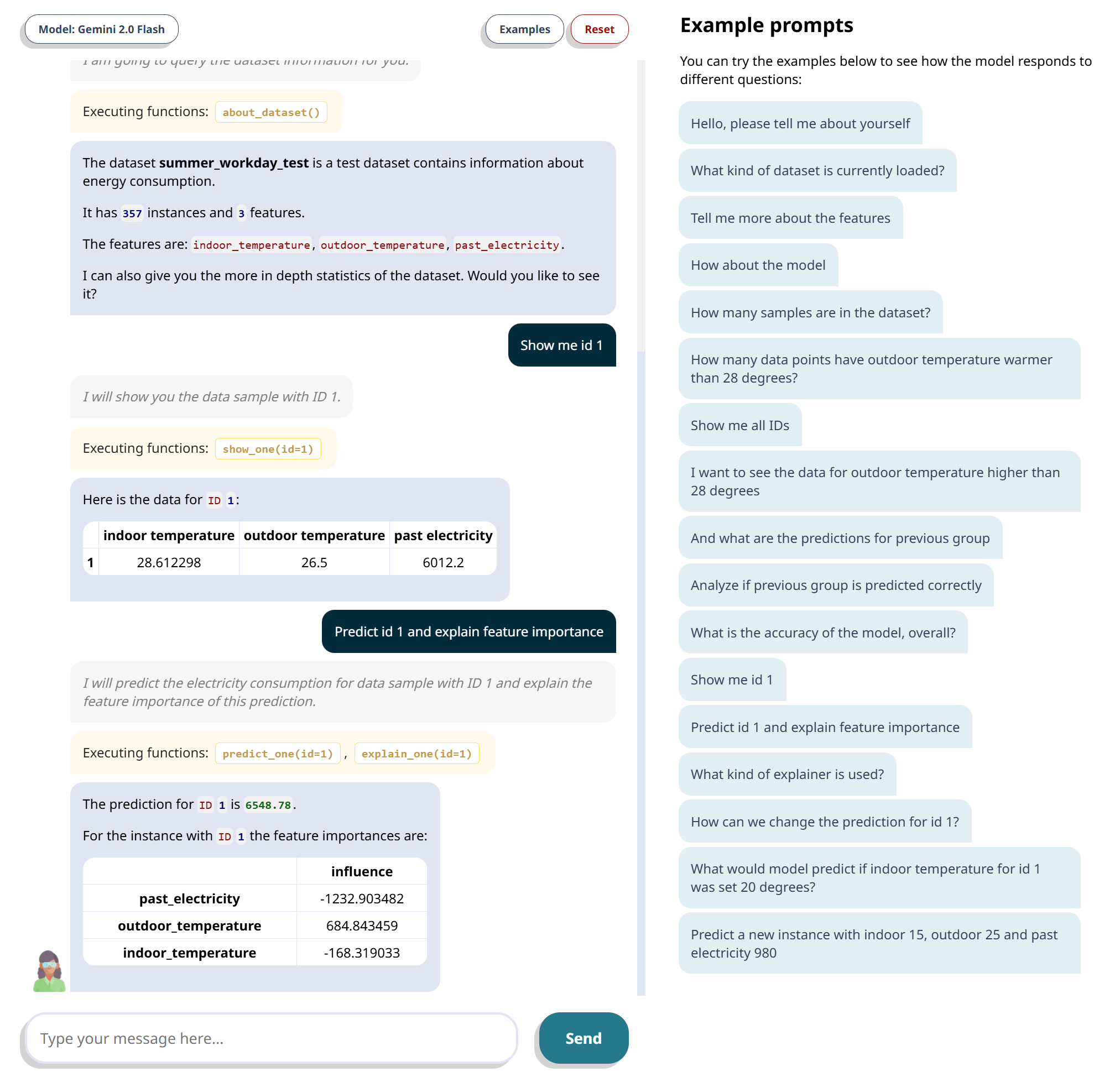}
    \caption{Explainability Assistant user interface.}
    \label{fig:ea-interface}
\end{figure*}

\textbf{Front-end.} The user interface is implemented using Next.js, providing a conversational interface for natural language interaction, as illustrated in Figure~\ref{fig:ea-interface}. The front-end manages client-side conversation history, handles user authentication and session management, and allows users to select from multiple supported LLMs (e.g., Llama-3.3-70B-Instruct, Gemini-2.0-Flash, Gemini-2.5-Flash). This design enables users to switch between models and providers, even within a single session.

\textbf{Back-end.} The server-side component is implemented as a FastAPI application that orchestrates data operations and ML model inference, executes pre-defined explainability functions, and manages communication with external LLM providers via API connections. The back-end leverages modern LLMs' function-calling capabilities: rather than fine-tuning models on task-specific grammars, the system provides the LLM with the user query, conversation history, and JSON specifications of available functions (including names, descriptions, and required parameters). The LLM then determines which function(s) to invoke and generates the appropriate arguments in a structured format.

This function-calling paradigm offers several key advantages:

\begin{itemize}
    \item \textbf{Improved Accuracy.} Intent extraction accuracy exceeds 90\% for capable models (e.g., Gemini-2.5-Flash achieves 93-94\%), compared to 76.8\% for the grammar-based system with T5 models in TalkToModel. This improvement is enabled both by the function-calling architecture and by advances in the underlying LLM capabilities.

    \item \textbf{Enhanced Transparency.} The LLM is constrained to respond using a structured JSON format that includes a mandatory free-form explanation field describing the assistant's planned actions (e.g., "I will predict the electricity consumption for sample ID 42 and explain which features most influenced this prediction"). This design exposes the system's intended actions and explicitly lists the functions to be executed, allowing users to verify the system's interpretation and making it more trustworthy.

    \item \textbf{Data Privacy.} Function execution occurs entirely within the back-end, retrieving results directly from local data and models without the necessity to transmit sensitive information to external LLM providers. Only the user query and tool schemas (and optionally summarized outputs) are sent to the LLM provider. This ensures that proprietary datasets and model internals can remain protected.

    \item \textbf{Deployment Flexibility.} The system adapts to new datasets and model types through configuration changes alone, as demonstrated with both energy consumption regression and clinical risk classification, without requiring task-specific fine-tuning.
\end{itemize}

It is worth noting that general-purpose tool-calling protocols, such as the Model Context Protocol (MCP)~\cite{anthropic2024mcp}, have recently emerged to standardize LLM-to-tool connectivity. However, the Explainability Assistant provides a domain-specific conversational layer that goes beyond raw tool invocation: it manages multi-turn XAI dialogues, performs automatic parameter selection for explanation methods based on conversational context, and presents results in a form accessible to non-expert users.

\subsection{XAI Tools}

The Explainability Assistant integrates several established post-hoc explanation techniques to support model interpretability.

\textbf{Feature Importance Analysis.} Understanding which features most influence model predictions is fundamental to model interpretability. The system employs SHAP~\cite{lundberg2017unified}, a model-agnostic method grounded in game-theoretic principles that provides consistent and theoretically justified feature attributions. Users can request explanations of feature importance for individual predictions (e.g., "Why was sample 42 predicted to have such consumption level?") or global summaries across the entire dataset (e.g., "What are the three most important features overall?").

\textbf{Counterfactual Explanations.} Counterfactual reasoning addresses contrastive questions of the form "What would need to change for the prediction to be different?" The system implements counterfactual generation using DiCE~\cite{mothilal2020explaining}, which identifies minimal, realistic modifications to input features that would alter the model's prediction. This capability supports actionable insights: for instance, in the energy consumption prediction use case, a specialist might ask, "What changes would move this prediction from high consumption to low?" to explore minimal required adjustments.

\textbf{What-If Scenario Analysis.} Beyond automated counterfactual generation, the system supports user-directed what-if queries that simulate specific hypothetical changes. Users can specify precise feature modifications and observe their effects on predictions, enabling exploratory analysis such as "If indoor temperature increases by 5 degrees, how much will energy consumption change?" This functionality is particularly valuable for policy analysis and sensitivity testing.

\subsection{Use Cases}

We demonstrate the capabilities of the Explainability Assistant through two distinct use cases: energy demand forecasting and heart disease classification.

\textbf{Energy Demand Forecasting (Regression).} This use case addresses mid-term forecasting of electricity demand in the commercial sector. The dataset originates from a real-world monitoring system deployed in an office building~\cite{sakkas2023forecasting}, combining smart meter readings with environmental sensor data. The primary predictors include indoor temperature, outdoor temperature, and time-shifted previous energy consumption values. We applied the EA to a Genetic Programming symbolic regressor model trained by the original stakeholder, enabling users to evaluate model performance and analyze feature attributions through natural language dialogue.

\textbf{Heart Disease Classification (Classification).} To demonstrate the architectural modularity of the Explainability Assistant beyond regression tasks, we deployed the system on the Heart Disease Cleveland dataset~\cite{heartcleveland}\footnote{\url{https://www.kaggle.com/datasets/ritwikb3/heart-disease-cleveland}} (303 patients, 14 clinical attributes, Decision Tree classifier). By simply updating the JSON function specifications, without modifying the core application logic or fine-tuning the LLMs, the system successfully transitioned from predicting continuous energy values to explaining binary classification risk factors. This confirms that the function-calling paradigm enables rapid adaptation to new domains.

\begin{figure*}[tbp]
    \centering
    \includegraphics[width=0.95\linewidth]{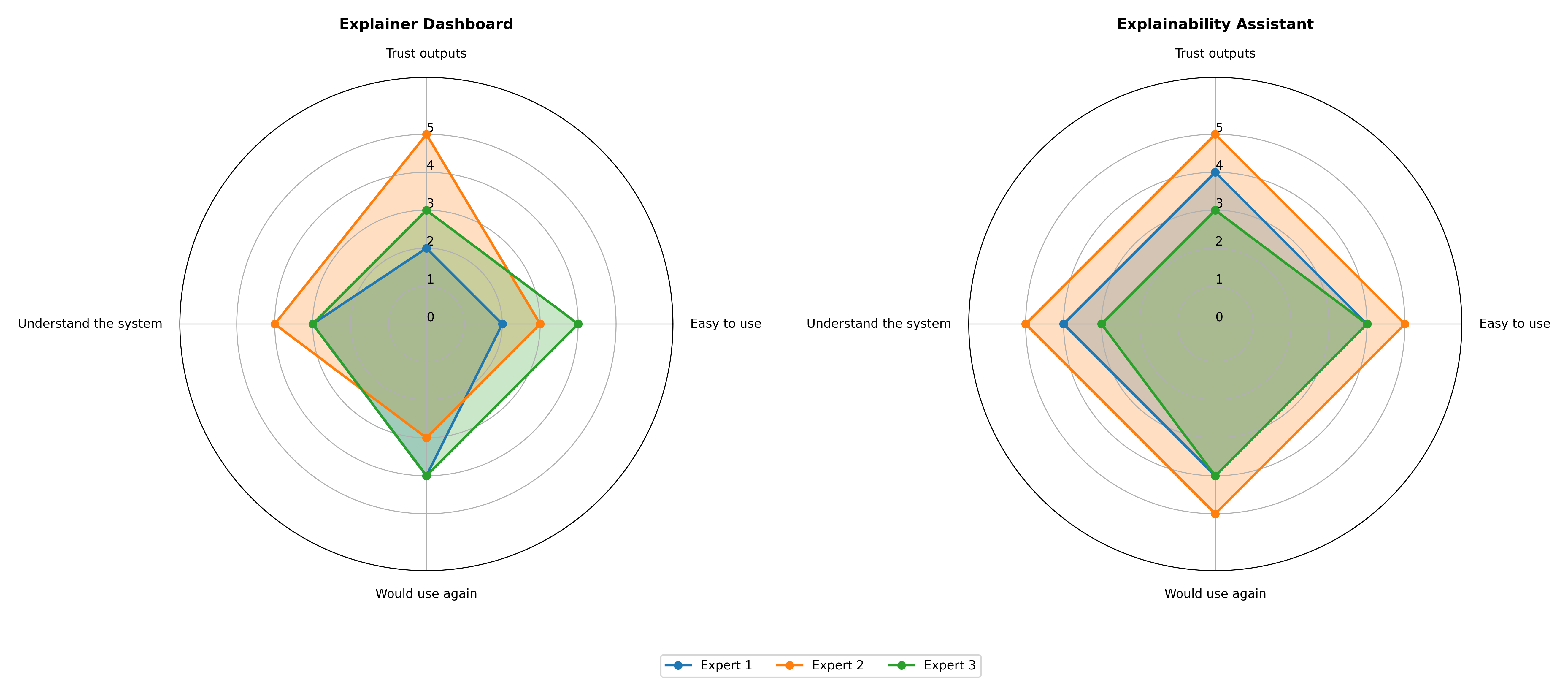}
    \caption{Individual expert ratings (1-5 Likert scale) comparing Explainer Dashboard (left) and Explainability Assistant (right) across four dimensions: trust in outputs, ease of use, system understanding, and likelihood of future use. Each colored line represents one expert's ratings.}
    \label{fig:radar}
\end{figure*}

\section{Evaluation}
\label{sec:evaluation}

The Explainability Assistant is evaluated through two complementary approaches: LLM parsing accuracy and an expert validation with energy domain specialists.

\subsection{LLM Parsing Accuracy}

Intent classification reliability was measured across multiple state-of-the-art LLMs using gold-standard parse datasets. Each evaluation instance consisted of user input, prior conversation history, and expected function call(s), as exemplified in Table~\ref{tab:sample}.

\begin{table}[ht]
\small
    \centering
    \begin{tabular}{|l|}
    \hline
    \textbf{User input:} \\
    "What is the prediction and how to change it?" \\
    \hline
    \textbf{Conversation history:} \\
    {[}"Show me the data", "Show id 38"{]} \\
    \hline
    \textbf{Expected function calls:} \\
    {[}"predict\_one(id=38)", "cfes\_one(id=38)"{]} \\
    \hline
    \end{tabular}
    \vspace{8pt}
    \caption{Gold parse dataset sample example.}
    \label{tab:sample}
\end{table}

Three gold parse datasets were constructed for the energy consumption forecasting use case. Dataset A (20 samples) was created manually to ensure coverage of core system functionalities (20 functions available in total). Datasets B and C (80 samples each) were generated using frontier LLMs GPT-5 and Gemini-2.5-Pro with Dataset~A as few-shot examples, then manually verified and corrected. Datasets~B and C were generated using LLMs that partially overlap with those evaluated, but manual verification of all samples and cross-validation on two combined sets ensure that the evaluation remains independent of the generation process.

Four cost-effective frontier LLMs were evaluated: GPT-5-mini, Llama-3.3-70B-Instruct, Gemini-2.0-Flash, and Gemini-2.5-Flash. Exact match parsing accuracy was measured on combined datasets A+B and A+C (100 samples each).

Table~\ref{tab:parsing} presents the results. Gemini-2.5-Flash achieved the highest accuracy (94\% on A+B, 93\% on A+C), followed by Gemini-2.0-Flash (88\% and 91\%). This demonstrates that modern function-calling architectures can achieve production-grade reliability (>90\%) without the task-specific fine-tuning required by earlier grammar-based systems, which typically plateaued around 75\% \cite{slack2023talktomodel}.

\begin{table}[ht]
\centering
\small
\begin{tabular}{|l|c|c|}
\hline
\textbf{Model} & \textbf{A+B} & \textbf{A+C} \\
\hline
GPT-5-mini & 74\% & 79\% \\
Llama-3.3-70B & 78\% & 74\% \\
Gemini-2.0-Flash & 88\% & 91\% \\
Gemini-2.5-Flash & \textbf{94\%} & \textbf{93\%} \\
\hline
\end{tabular}
\vspace{8pt}
\caption{Exact match parsing accuracy on combined gold parse datasets.}
\label{tab:parsing}
\end{table}

\subsection{Expert Validation Design}

To assess practical utility, we conducted a within-subjects comparison study with three energy domain specialists. Participants were professionals actively involved in building energy management and policy decisions, with 10-20 years of experience in the energy sector and varying levels of familiarity with ML systems (self-reported as intermediate to expert). This expert-driven evaluation prioritizes depth of domain insight over sample breadth, consistent with established practices for evaluating specialized decision-support tools in engineering contexts. The small participant count reflects the limited pool of domain experts with direct access to the deployed building energy management system and familiarity with its underlying ML models; the within-subjects design partially compensates for the small sample size by ensuring that individual differences are controlled across conditions.

Participants engaged in a simulated decision-support scenario: managing energy consumption in a commercial building during business hours. An ML model provided hourly energy consumption predictions (regression output) based on three features: indoor temperature, outdoor temperature, and past consumption. We compared two interfaces providing access to the same underlying model and explanation methods:

\textbf{Explainer Dashboard (ED):} A traditional point-and-click graphical interface based on the open-source Explainer Dashboard framework\footnote{\url{https://explainerdashboard.readthedocs.io}}. The dashboard provided tabbed access to SHAP feature importances, individual predictions, global feature importance rankings, and interactive sliders for what-if analysis.

\textbf{Explainability Assistant (EA):} The conversational interface described in Section~\ref{sec:system}, configured with the same dataset and model. Users interacted through natural language queries to request predictions, explanations, and scenario analyses.

The study was conducted asynchronously via Google Forms to accommodate participants' schedules. Each participant received links to both interface deployments along with detailed usage instructions. The survey consisted of six sections: (1) background information collection, (2) Task Block A using ED, (3) subjective evaluation of ED, (4) Task Block B using EA, (5) subjective evaluation of EA, and (6) open-ended feedback.

Each task block contained five questions requiring participants to synthesize ML predictions, XAI explanations, and domain knowledge. Questions were carefully designed to be parallel in structure and difficulty while using different data instances to prevent memorization. Example tasks included: "What is the predicted energy consumption for hour X?", "Which feature most influences this prediction?", "What would happen to consumption if indoor temperature increased by Y degrees?", "How accurate is the model overall?", and "Identify a case where the model makes an error." Complete task questions are provided in Appendix~\ref{sec:appendix-study}.

Following each task block, participants rated four dimensions on five-point Likert scales (1=Strongly Disagree, 5=Strongly Agree): ease of use, trust and confidence in the interface, understanding of how the system works, and likelihood of using the interface in future work.

Our design deliberately emphasized measuring appropriate reliance and objective task accuracy rather than subjective trust alone, addressing concerns about potential illusion of explanatory depth in conversational systems~\cite{he2025conversational}.

\subsection{Expert Validation Results}

The participants completed ten task questions (five per interface) with high accuracy, achieving 93\% correct responses using ED and 100\% using EA. Although not statistically significant given the sample size, this suggests that the conversational interface maintains or improves objective performance while offering usability advantages.

Figure~\ref{fig:radar} presents individual expert ratings across four evaluation dimensions. For the Explainer Dashboard (left), ratings varied considerably: Expert~1 (blue) provided low ratings (2-3), Expert~2 (orange) gave the highest ratings (3-5), and Expert~3 (green) showed intermediate ratings.

The Explainability Assistant (right) received consistently higher and more uniform ratings. All three experts rated "Easy to use" and "Would use again" at or near maximum (5), demonstrating unanimous agreement on usability and practical value. Experts 1 and 3 showed slightly more measured enthusiasm for "Trust outputs" and "Understand system" (3-4), while Expert 2 provided near-perfect scores across all dimensions.

The most striking contrast appears in "Would use again," where EA achieved unanimous maximum ratings (5) compared to ED's variable scores (2-4). Similarly, "Easy to use" improved dramatically from ED (2-5, with disagreement) to EA (all 5s). These patterns suggest the conversational paradigm successfully addresses usability barriers in traditional dashboards.

The more conservative ratings for "Trust outputs" and "Understand system" (3-5) suggest participants maintained appropriate epistemic caution about model outputs and system mechanics, aligning with the design goal of supporting appropriate reliance rather than uncritical trust. Participants noted that ED's visual overview was occasionally useful for orientation, suggesting hybrid approaches may offer complementary benefits.

\subsection{Discussion} \label{discussion}

The evaluation demonstrates that the Explainability Assistant achieves two key
objectives: (1) substantially improved reliability in comprehension compared to prior conversational XAI systems, and (2) offered practical utility for domain experts, allowing actionable insights based on ML model predictions. The 94\% intent classification accuracy, achieved through effective integration of LLM function-calling with frontier models, addresses a central limitation of earlier grammar-based systems, where frequent
misinterpretations undermined user's trust and prevented production deployments. The structured function-calling format further increases transparency by exposing the system's intended actions before
execution.

The expert validation shows that the conversational interface reduces usability
friction without sacrificing analytical quality. Experts unanimously preferred
EA in terms of usability and future adoption while maintaining high task
accuracy. At the same time, ratings for "trust outputs" and "understand
system" remained moderate, suggesting appropriate epistemic caution, which is aligned with our
design goals. Participants also noted that traditional dashboards still provide
value for global orientation, indicating that hybrid conversational-visual
interfaces merit exploration.

\section{Conclusion}
\label{sec:conclusion}

We presented the Explainability Assistant, a conversational XAI system that leverages LLM function-calling to achieve 94\% intent classification accuracy, representing a substantial improvement over prior grammar-based systems. The modular architecture eliminates task-specific fine-tuning requirements and enhances transparency through structured responses that expose system reasoning before execution.

Our evaluation with energy domain specialists demonstrated that the conversational paradigm successfully addresses usability barriers in traditional dashboards, with all three experts unanimously preferring the Explainability Assistant for ease of use and likelihood of future adoption while maintaining high task accuracy (100\% vs. 93\%). The system advances conversational XAI through improved comprehension reliability, reduced deployment complexity, enhanced user experience, and reasoning transparency. By making sophisticated XAI techniques accessible through natural language, the system lowers barriers for domain experts to engage with ML model understanding regardless of their technical background. The open-source implementation enables the research community to extend conversational XAI to new application domains.

Future work should include larger-scale studies with diverse user populations across multiple domains, temporal analyses of extended usage patterns to understand long-term adoption, and investigation of hybrid interfaces that combine conversational flexibility with strategic use of visualization, as suggested by participant feedback.

\section*{ACKNOWLEDGMENT}
This research was conducted under the Transparent, Reliable, and Unbiased Smart Tool for AI (Trust-AI) project (Grant Agreement No.~952060), funded by the European Commission. This work was also supported by the Estonian Research Council grant PRG2006.

\section*{Review Materials}

To facilitate evaluation, readers are provided with: (1) a video demonstration
of the system, (2) a public demo interface, (3) the complete source code,
(4) gold parse datasets and evaluation scripts, and (5) the full expert validation
questionnaire. All links are listed in Appendix~\ref{sec:resources}.

\appendix

\section{Ethics Statement}

This work follows established research ethics guidelines. The Explainability Assistant is designed to improve transparency and appropriate reliance in machine learning systems by exposing the model's intended actions before execution and avoiding hidden
reasoning steps. All data used in system development are either synthetic, publicly available (e.g., Heart Disease Cleveland dataset), or provided by industry partners in anonymized form. No personally identifiable information is processed or transmitted to LLM providers when all function execution and data
access occur locally within the back-end. The expert validation involved domain experts using non-sensitive energy data; no personal information or behavioral logs were collected. The system is intended for decision support and not for autonomous or high-stakes decision-making.

\section{Resources}
\label{sec:resources}

\begin{enumerate}
    \item Explainability Assistant website: \\ \href{https://explainabilityassistant.com}{https://explainabilityassistant.com}

    \item Demo access to the EA chat interface: \\ \href{https://explainabilityassistant.com/demo}{https://explainabilityassistant.com/demo}

    \item Introductory video presentation about Explainability Assistant: \\ \href{https://www.youtube.com/watch?v=36Tm3kof-tc}{https://www.youtube.com/watch?v=36Tm3kof-tc}

    \item Explainability Assistant, back-end application code: \\ \href{https://github.com/krkv/explainability-backend}{https://github.com/krkv/explainability-backend}

    \item Explainability Assistant, front-end application code: \\ \href{https://github.com/krkv/explainability-frontend}{https://github.com/krkv/explainability-frontend}

    \item Gold parse datasets, dataset generation prompts and scripts, LLM evaluation scripts: \\ \href{https://github.com/krkv/llm-evaluation}{https://github.com/krkv/llm-evaluation}
\end{enumerate}

\section{Expert Validation Questions}
\label{sec:appendix-study}

\subsection{Introduction and Background Questions}

Hello! This survey is designed to evaluate your experience and compare preferences in using two different software tools for explainable machine learning.

In this scenario, a machine learning model is used to forecast energy consumption of a building based on time series data (indoor temperature, outdoor temperature and energy consumption in the past hour).

You will need to find answers to a few questions about these predictions. It should take you up to 30 minutes. Please read the instructions in the beginning of every section. Thank you for participating!

\begin{itemize}
    \item How many years of experience do you have in the energy domain?
    \item How familiar are you with machine learning algorithms and applications?
    \item What is your current role? (optional)
\end{itemize}

\subsection{Block A Task Questions}
For answering these questions, please open the Explainer Dashboard app and find the answers there. If you can't find the answer, leave the field empty or select "I can't tell".

\begin{itemize}
    \item What is the prediction for ID 33?
    \item Which feature contributed the most to this prediction?
    \item How would the prediction for ID 33 change if the indoor temperature changed to 20 degrees?
    \item How big is the error in the prediction of ID 33?
    \item Find another example of how the model prediction can change (what are the input values and what is the model prediction)
\end{itemize}

\subsection{Block A Post-Task Questions}

Please answer if you agree with these statements about your experience using the Explainer Dashboard app.

\begin{itemize}
    \item It was easy to use the dashboard to find the answers.
    \item I trust the dashboard and the outputs it gave.
    \item I understand how this system operates.
    \item I would use this type of tool again in the future.
\end{itemize}

\subsection{Block B Task Questions}

For answering these questions, please open the Explainability Assistant app and find the answers there. If you can't find the answer, leave the field empty or select "I can't tell".

\begin{itemize}
    \item What is the prediction for ID 55?
    \item Which feature contributed the most to this prediction?
    \item How would the prediction for ID 55 change if the indoor temperature changed to 20 degrees?
    \item How big is the error in the prediction of ID 55?
    \item Find another example of how the model prediction can change (what are the input values and what is the model prediction)
\end{itemize}

\subsection{Block B Post-Task Questions}

Please answer if you agree with these statements about your experience using the Explainability Assistant app.

\begin{itemize}
    \item It was easy to use the assistant to find the answers.
    \item I trust the assistant and the outputs it gave.
    \item I understand how this system operates.
    \item I would use this type of tool again in the future.
\end{itemize}

\subsection{Open-ended Feedback}

Feel free to go back to the Explainability Assistant tool and explore it more. You can try different types of questions, also there are examples that can be toggled in the top right corner.

Please share any feedback, ideas and thoughts you might have after completing these questions and interacting with both interfaces.

Your opinion as an expert is very important, and we are very thankful for your contribution!

\section{Limitations}

Several limitations of this work warrant consideration. The expert validation involved a small sample size (n=3), which limits statistical power and generalizability of the findings. However, the within-subjects design and unanimous preference patterns on key usability dimensions provide meaningful initial evidence of the system's practical utility.

The asynchronous study format, while accommodating participants' schedules, prevented observation of real-time interaction patterns and think-aloud protocols that could reveal deeper insights into how experts leverage each interface and formulate queries during exploratory analysis.

Because the study used a fixed interface order (Explainer Dashboard first, then Explainability Assistant), participants may have benefited from learning/familiarization with the task structure and data in Block A, potentially inflating performance or usability ratings in Block B. Future work should counterbalance interface order to control for these ordering effects.

Our evaluation focused on a single domain (energy forecasting) with a specific model type (symbolic regression). While the architectural modularity and demonstrated parsing accuracy across diverse query types suggest potential for broader applicability, validation in other domains, especially in high-stakes settings like healthcare, remains necessary to establish generalizability.

\bibliographystyle{IEEEtran}
\bibliography{ea}

\end{document}